\documentclass[10pt,journal,compsoc]{IEEEtran}
\usepackage{multicol}
\usepackage{multirow}
\usepackage{booktabs}
\usepackage{threeparttable}
\usepackage{array}
\usepackage{booktabs}
\usepackage{bbding}
\usepackage{diagbox}
 \usepackage{graphicx}
 \usepackage{amsmath}
 \usepackage{amsmath}
  \usepackage{amsfonts}
  \usepackage[mathscr]{euscript}
\usepackage{hyperref}
\usepackage[mathscr]{euscript}
\usepackage{booktabs}       
\usepackage{graphicx}
\usepackage{subfigure}
\ifCLASSOPTIONcompsoc
  \usepackage[nocompress]{cite}
\else
  \usepackage{cite}
\fi
\ifCLASSINFOpdf
\else
\fi
\begin{document}
%
\title{Domain-Unified Prompt Representations for Source-Free Domain Generalization}
%
%
%
%

\author{Hongjing Niu$^{\ast}$, Hanting Li$^{\ast}$, Feng~Zhao$^{\dagger}$,~\IEEEmembership{Member,~IEEE},
\and Bin~Li,~\IEEEmembership{Member,~IEEE}\\
 \thanks{$^{\ast}$ Equal contribution}  \thanks{$^{\dagger}$ Corresponding to fzhao956@ustc.edu.cn} \thanks{H. Niu, H. Li, F. Zhao, and B. Li are affiliated with University of Science and Technology of China, Hefei 230026, China.}}
\IEEEtitleabstractindextext{%
\begin{abstract}
Domain generalization (DG), aiming to make models work on unseen domains, is a surefire way toward general artificial intelligence. Limited by the scale and diversity of current DG datasets, it is difficult for existing methods to scale to diverse domains in open-world scenarios (e.g., science fiction and pixelate style). Therefore, the source-free domain generalization (SFDG) task is necessary and challenging. To address this issue, we propose an approach based on large-scale vision-language pretraining models (e.g., CLIP), which exploits the extensive domain information embedded in it. The proposed scheme generates diverse prompts from a domain bank that contains many more diverse domains than existing DG datasets. Furthermore, our method yields domain-unified representations from these prompts, thus being able to cope with samples from open-world domains. Extensive experiments on mainstream DG datasets, namely PACS, VLCS, OfficeHome, and DomainNet, show that the proposed method achieves competitive performance compared to state-of-the-art (SOTA) DG methods that require source domain data for training. Besides, we collect a small datasets consists of two domains to evaluate the open-world domain generalization ability of the proposed method. The source code and the dataset will be made publicly available at \href{https://github.com/muse1998/Source-Free-Domain-Generalization}{https://github.com/muse1998/Source-Free-Domain-Generalization}
\end{abstract}
\begin{IEEEkeywords}
Dynamic facial expression recognition, expression intensity, deep learning.
\end{IEEEkeywords}
}
\maketitle
\IEEEdisplaynontitleabstractindextext

%
\IEEEpeerreviewmaketitle

\IEEEraisesectionheading{
\section{Introduction}}

Deep learning \cite{DeepLearning, ResNet} has achieved great success in many fields, especially computer vision.
However, the generalization of deep learning methods is not yet good enough, which has become a constraint for wide applications.
In recent years, deep learning generalizability has attracted increasing attentions and sprouted several subfields. Although these studies differ somewhat in their task settings, their ultimate goals are the same, i.e., to enabling models to cope with a wide range of possible scenarios in the open world.

Generalizability has long been an essential topic in the field of machine learning.
In practical application scenarios, the ability to make reasonable predictions on unseen data is critical for machine learning models.
In recent years, thanks to the advancement of algorithms and the increase in data scale, deep learning models have achieved satisfactory performance in testing scenarios that are independently and identically distributed with the training dataset.
Furthermore, deep learning models could face more challenges, where the testing and training data may obey different distributions.

Domain generalization (DG) is one of the most representative tasks and has attracted considerable attentions.
For image classification, the DG tasks provide datasets from multiple domains that may come from different acquisition methods or have different image styles. These data are divided into a source domain (for training) and a target domain (for testing), without overlapping between them.
For example, models are trained on a photo dataset and tested on a sketch dataset.

For DG tasks, the source domain data are often an important factor indicating the difficulty of the task.
When the source domain data is rich, the model learns domain-invariant features more easily and presents better generalization capabilities.
In the case of PACS \cite{PACS} dataset, for example, an approach using three source domains can achieve nearly double performance of the one that utilizes a single source domain.
In real-world tasks, obtaining sufficiently rich source domain data is often not guaranteed. Reducing the dependence on source domain data in domain generalization is a widely applied and exciting problem, so we propose the source-free domain generalization (SFDG) task. The model is used directly to predict the target domain samples without using the source domain data for training.

To solve the SFDG problem, we apply a vision-language model for feature extraction.
Compared to acquiring a huge amount of images as source domain data, representing the source domain features in the form of text is almost costless.
After obtaining the textual encodings of different domains by constructing various textual prompts, we propse a domain-unified prompt representation generator (DUPRG) to integrate these text encodings and aggregate a set of domain-unified text representations.
For the target domain samples from the open world, the model prediction is made based on the domain-unified text representations.
With the vision-language model, we only need to process the text encoding to obtain domain invariance in the training phase, thus bypassing the need for a large amount of source domain data and achieving source-free domain generalization.

Unlike most DG methods, our model does not require source domain data for training, which greatly extends the application scenarios of the task.

The main contributions of this work can be summarized as follows.
\begin{itemize}
\item The research problem of SFDG is introduced, which is more suitable for most application scenarios than the basic DG task.
\item An effective SFDG method is proposed to achieve DG for visual tasks by learning domain-unified text encodings. The proposed framework has comparable or even better performance than other DG approaches using source domain training on mainstream datasets.
\end{itemize}

\section{Related Work}\label{sec:related}
\subsection{Domain Generalization}
In this paper, a more challenging task, named source-free domain generalization is proposed based on the domain generalization task.
In computer vision, a seminal work \cite{Torralba2011CVPR} suggests that dataset biases can lead to poor generalization performance.
For example, a person classifier trained on Caltech101 \cite{Cal101} could obtain a very low accuracy (11.8\%) on LabelMe \cite{LabelMe}.
The domain generalization issue raised attentions, and many wonderful studies have tried to address it since then.
From a methodological point of view, the DG approaches can be broadly divided into domain alignment \cite{Muandet2013ICML, Li_2018_ECCV}, meta-learning \cite{Balaji2018MetaReg}, data augmentation \cite{Zhou2020ECCV}, ensemble learning \cite{Liu2020MSNET}, self-supervised learning \cite{Carlucci_2019_CVPR}, disentangling representations \cite{khosla2012undoing}, regularization strategies \cite{wang2018learning}, etc., which has been discussed in detail in a survey \cite{zhou2022domain}.

Here, we discuss some nuances of the existing DG tasks from a task setting perspective.
In the basic DG task, the dataset consists of samples from multiple domains \cite{PACS, VLCS, OfficeHome, Terra, DomainNet}, and in addition to the category labels, domain labels are also provided. These samples are divided into source and target domains, where the source domain is used for training and validation, while the target domain is used for testing only.
Domainbed \cite{gulrajani2021in} provides a complete process, and many methods follow its experimental settings.
Single-source DG \cite{Wang_2021_ICCV} is more challenging, since its source domain has only one domain.
This makes it difficult to find common domain features. Therefore, most approaches opt for less affected strategies such as data augmentation.
Some attempts, like unsupervised DG \cite{Zhang_2022_CVPR}, are made to improve the generalizability from a pretraining perspective. By replacing the commonly used ImageNet pretraining with an unsupervised pretraining approach, the model achieves better performance on downstream DG tasks.
Zero-shot DG \cite{maniyar2020zero} modifies the category space of the target domain so that it no longer coincides with the category space of the source domain.

It can be seen that various difficult versions of previous DG tasks, including unsupervised DG and zero-shot DG, cannot completely discard the source domain.

\subsection{Vision-language pretraining}
In contrast to natural images, language, as a symbolic representation created by humans, is inherently rich in prior knowledge and well interpretable, which inspires researchers to exploit the rich semantic information contained in natural language to help the neural networks to learn a better visual representation  \cite{su2019vl,chen2020uniter,li2020unicoder,tan2019lxmert,lu2019vilbert}. These methods effectively enhance the performance of many
cross-modal tasks, such as visual question answering (VQA) \cite{antol2015vqa} and visual commonsense reasoning (VCR) \cite{zellers2019recognition}.

Specially, contrastive language-image pretraining (CLIP) \cite{radford2021learning} utilizes 400 million (image, text) pairs collected from the internet to learn robust and superior image representations through contrastive learning. Recently, some methods find that CLIP is exceptional at encoding the semantic meaning of visual depictions, regardless of their styles \cite{vinker2022clipasso,goh2021multimodal}, which is in line with the original purpose of the domain generalization task, i.e., learning the uniform visual semantic representations across domains. Therefore, \cite{li2022domain} introduced a novel domain generalization paradigm to better leverage various large-scale pretraining models, including CLIP. \cite{zhang2021amortized} devised domain prompt learning (DPL) to generate a domain-specific prompt for each image. Similarly,  Zheng et al. trained a prompt adapter to produce a suitable prompt for each input image. \cite{cha2022domain} derived a tractable variational lower bound via approximating the oracle model by a pretrained model, termed as mutual information regularization with oracle (MIRO).

The above approaches require more or less training on the source domain to learn the uniform semantics between different domains. They suffer from two main problems: 1) \textit{Limited source domain data is hard to be extended to the rich unknown domains in open-world scenarios (e.g., science fiction and pixelate style)}, and 2) \textit{Well-developed visual semantic and rich domain information embedded in large-scale vision-language pretraining models are not fully utilized}. Therefore, we propose a novel domain-unified prompt representation generator to deal with the source-free domain generalization task in open-world scenarios.

\section{Method}
\subsection{Problem Formulation}
The domain generalization task aims to address the shift in data distribution between different domains by zero-shot transferring knowledge from the source domain to the unseen domain. DG methods usually train a model $f: \mathbb{R}^{H\times W}\mapsto \mathbb{R}^{C}$ on several source domains $D_{s}=\{D_{1}, D_{2}, \cdot\cdot\cdot\, D_{S}\}$ and then evaluate it on unseen target domains $D_{t}=\{D_{1}, D_{2}, \cdot\cdot\cdot\, D_{T}\}$. $D_{K}=\{(x_i,y_i)\}_{i=1}^{N}$ represents a dataset of the domain $K$, which samples $N$ sample points from an input image space $\mathcal{X}_{K} \in \mathbb{R}^{H\times W}$ and a domain-shared label space $\mathcal{Y} \in \mathbb{R}^{C}$, where $H$ and $W$ are the height and width of the input images, and $C$ is the number of the classes.

However, the existing domain generalization datasets (e.g., PACS \cite{li2017deeper}, and DomainNet \cite{peng2019moment}) contain only a very limited number of domains (e.g., painting, sketch, and cartoon), which makes it difficult for networks trained on these datasets to generalize to unseen domains (e.g., mosaic and abstract) in the open-world scenarios. Therefore we first introduce a source-free domain generalization task that aims to generalize the visual representations learned by the neural networks to any unseen domain of real-world scenarios without training through the source domain data $D_{s}$ (i.e., without sampling images from the source domain).

\begin{figure*}[t]
  \centering
  \includegraphics[width=\linewidth]{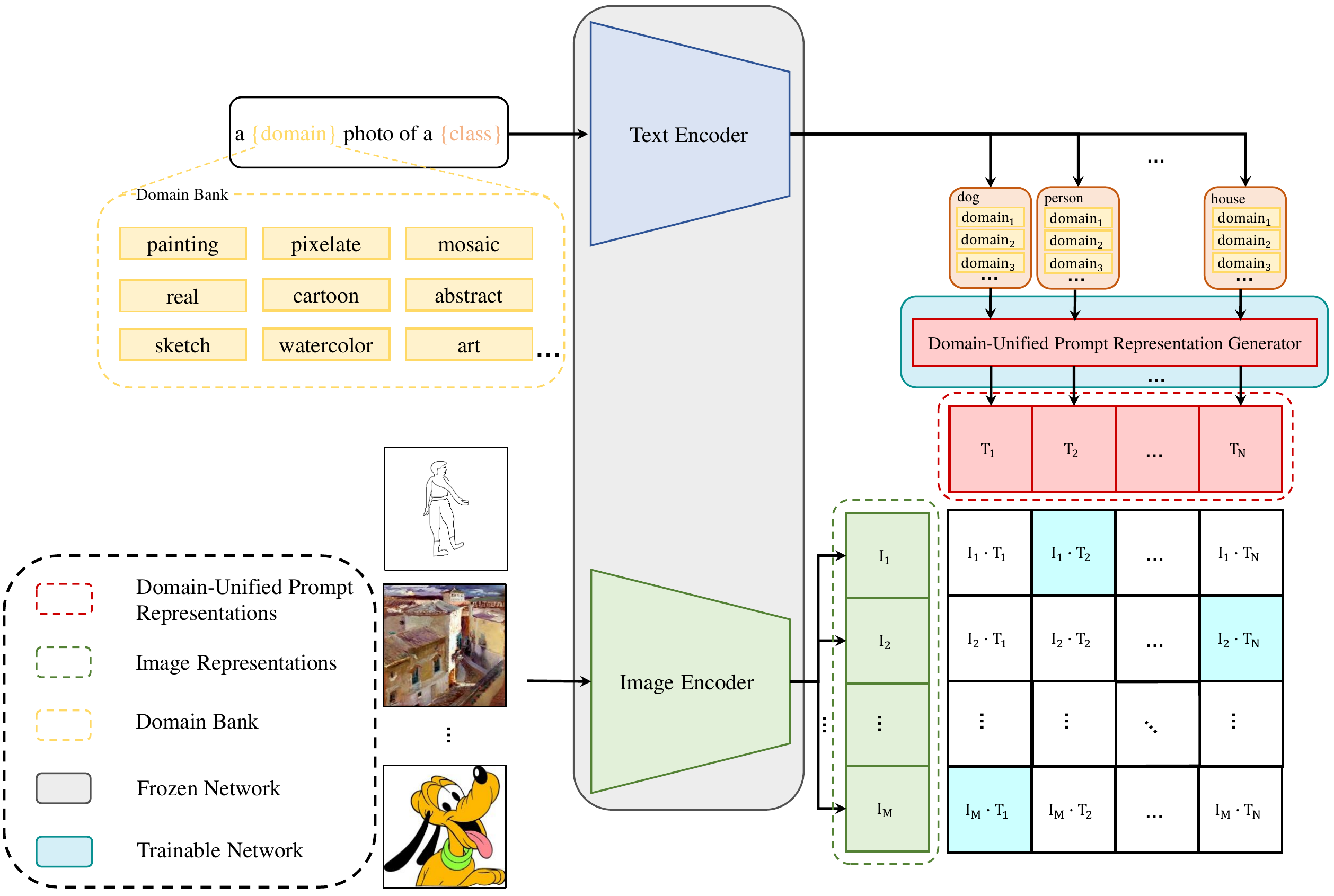}
  \caption{Overview of our method for SFDG. First, based on the rich variety of domains in domain bank (DB), we generate $M$ text descriptions under different domains for each class, where $M$ is the size of the domain bank. After that, our proposed DUPRG generates a domain-unified prompt representation for each class for inference. It is worth noting that only DUPRG needs to be trained in the whole paradigm while the parameters of both text and image encoders are frozen. }
  \label{fig:pipeline}
\end{figure*}
\subsection{CLIP for source-free domain generalization task}
To achieve generalization in open-world scenarios, our model needs to consistently model different semantics under a rich set of domains. However, it is not easy to encompass the domain knowledge in the real world using only the current limited-scale DG datasets.

Recently, large-scale vision-language pretraining models \cite{radford2021learning,li2022blip} obtain uniform and robust semantic representations \cite{goh2021multimodal} by contrast learning over a large number of (image, text) pairs. Especially, CLIP match the accuracy of the original ResNet50 \cite{he2016deep} on ImageNet \cite{deng2009imagenet} zero-shot without needing to use any images from ImageNet. We believe that the extensive domain information embedded in CLIP is the key to solving the SFDG task.

As a vanilla approach, CLIP uses its text encoder and image encoder to embed the input prompts (``a photo of a \textbf{\{class\}}'') and images into text features $T_{i} \in \mathbb{R}^{d}$ and image features $I_j \in \mathbb{R}^{d}$ of the same dimension $d$, respectively. Then CLIP gives the inference result by calculating the cosine similarity of $T_{i}$ and $I_{j}$, which can be formulated as,

\begin{equation}\label{1}
\widehat{y}_j =\mathop{\arg\max}\limits_{i}(\langle I_{j}, T_{i} \rangle),\quad\quad\quad\quad\quad i\in \left\{1,2,\cdot\cdot\cdot ,C \right \}.
\end{equation}

\noindent Where $\widehat{y}_j$ is the prediction made by CLIP on the $j$-th image, $C$ is the number of the classes, and $\langle\cdot, \cdot\rangle$ represents the cosine similarity between two vectors.
For SFDG tasks, a naive strategy utilizing CLIP is to use a standard prompt to generalize text prompts for each category, and then finish the prediction following the above inference process.
\begin{figure*}[t]
  \centering
  \includegraphics[width=\linewidth]{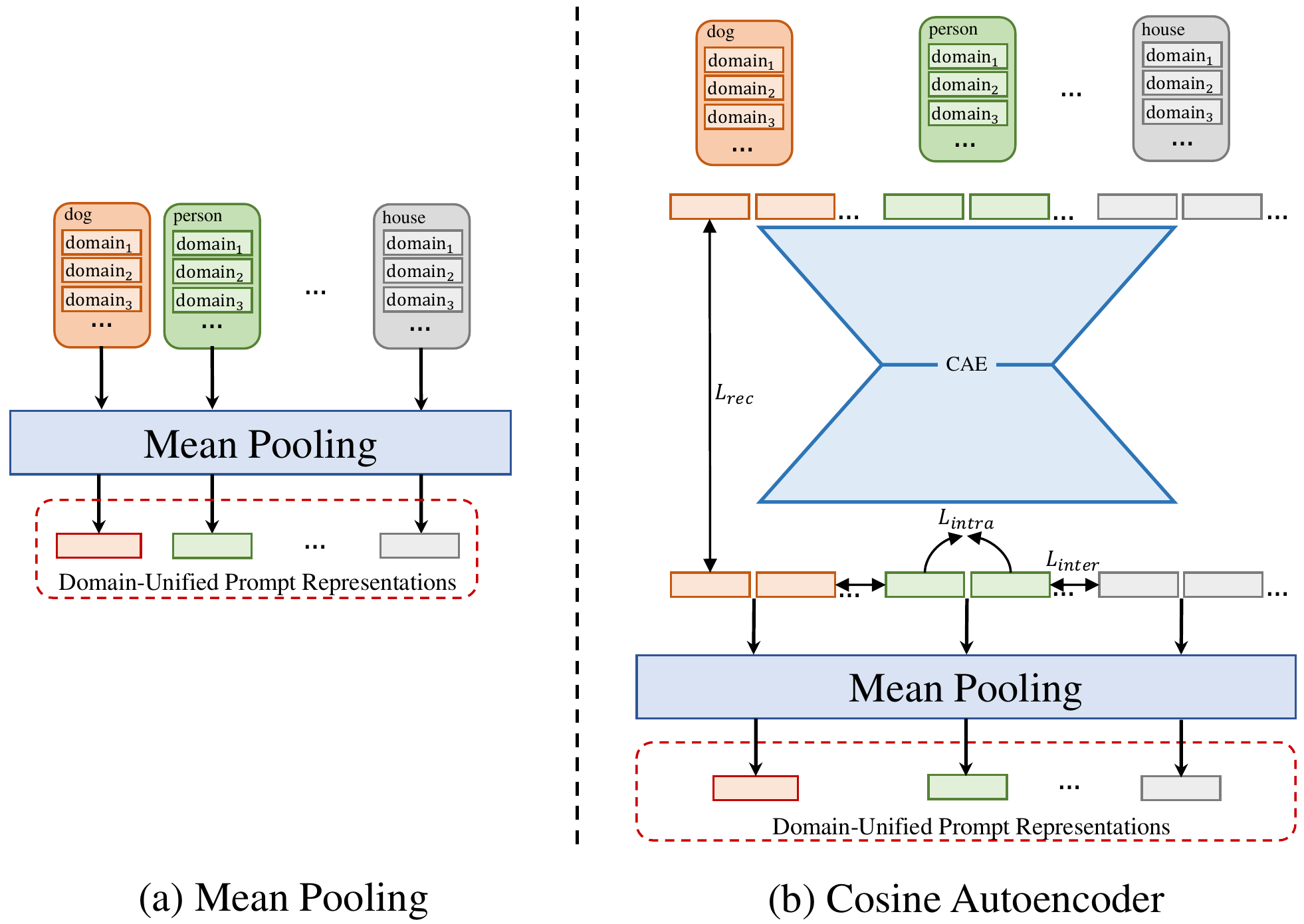}
  \caption{Two implementations of the proposed domain-unified prompt representation generator.}
  \label{fig:duprg}
\end{figure*}

\subsection{DUPRG: a new paradigm for source-free domain generalization task}
Although experiments in \cite{zhang2021amortized} show a good performance on several DG datasets with the standard prompt (``a photo of a \textbf{\{class\}}''). The standard prompt cannot provide an equally robust prompt for visual semantic representations of different open-world domains. For example, ``a \textbf{\{watercolor\}} photo of \textbf{\{class\}}'' can be a better prompt for watercolor images. To accomplish the SFDG task, we need to generate domain-unified prompt representations for each category to cope with the rich open-world domains.

Based on this viewpoint, we build a domain bank, which contains various domains that exist in the open-world scenarios. As shown in Fig.~\ref{fig:pipeline}, according to the domain type in the domain bank, $M$ prompts are generated for each class, which means that we can get $M\times C$ prompt representations embedded by the text encoder.

After getting the $M$ prompt representations of each class, these prompt representations are aggregated by the DUPRG into a domain-unified prompt representation for final inference. We propose two specific implementations of DUPRG in Fig.~\ref{fig:duprg}.

\textbf{Mean Pooling}: As shown in Fig.~\ref{fig:duprg}(a), the most intuitive way to aggregate these $M$ representations is to take the mean value as the domain-unified prompt representation. The domain-unified prompt representation of the $i$-th class can be defined as,
\begin{equation}\label{2}
T_i=\frac{\sum_{j=1}^{M}T_i^{j}}{M},\quad\quad\quad\quad\quad i\in \left\{1,2,\cdot\cdot\cdot ,C \right \},
\end{equation}
where $T_i^{j}$ denotes the prompt representation of the $i$-th class in the $j$-th domain.

\textbf{Cosine Autoencoder}: As shown in Fig.~\ref{fig:duprg}(b), we propose a cosine autoencoder (CAE) to extract the shared part of the original $M$ prompt representations (i.e., class semantic related part). A naive autoencoder computes the mean squared error (MSE) between the input and output as the reconstruction loss function \cite{AE}. However, the inference process of CLIP uses the cosine similarity to make the final prediction. Therefore, to preserve the semantically relevant orientation in the reconstruction process, we use cosine similarity as the reconstruction loss of the proposed CAE. The reconstruction loss can be formulated as,

\begin{equation}\label{3}
\mathscr{L}_{rec} = -\frac{\sum_{i=1}^{C}\sum_{j=1}^{M}\langle T_{i}^j, \widehat{T_{i}^j} \rangle}{MC},
\end{equation}
where $\widehat{T_{i}^j}$ is the reconstruction output of CAE for $T_{i}^j$. In addition, in order to obtain a domain-unified representation, an intra-class loss is used to narrow the gap between representations of different domains, which can be computed as,
\begin{equation}\label{4}
\mathscr{L}_{intra} = -\frac{\sum_{i=1}^{C}\sum_{j=1}^{M}\langle T_{i}^j, \overline{T_{i}} \rangle}{MC},
\end{equation}
where $\overline{T_{i}}$ is the mean value of $M$ prompt representations of $i$-th class. Finally, an inter-class loss is defined to make the cosine similarity of the representations of different classes in the same domain as low as possible (i.e., obtain a larger inter-class distance in each domain). The inter-class loss can be defined by,
\begin{equation}\label{5}
\begin{aligned}
\mathscr{L}_{inter} &= \frac{\sum_{i=1}^{M}\sum_{j=1}^{C}\sum_{k=1,\, j\ne k}^{C}\langle T_{j}^i, T_{k}^i \rangle}{MC(C-1)}.
\end{aligned}
\end{equation}

The overall loss function of CAE is defined as follows,
\begin{equation}\label{6}
\begin{aligned}
\mathscr{L}_{all} =\mathscr{L}_{rec}+ \lambda_1 \mathscr{L}_{intra}+ \lambda_2 \mathscr{L}_{inter}.
\end{aligned}
\end{equation}
Where $\lambda_1$ and $\lambda_2$ are two hyper-parameters controlling loss coefficients. The domain-unified prompt representation can be directly calculated by,

\begin{equation}\label{7}
T_i=\frac{\sum_{j=1}^{M}\widehat{T_i^{j}}}{M},\quad\quad\quad\quad\quad i\in \left\{1,2,\cdot\cdot\cdot ,C \right \},
\end{equation}
It is worth noting that no images are involved in the whole training process of CAE. Our method takes $M\times C$ prompt representations generated from the domain bank as the input to train the CAE. Since our method does not overfit the limited source domain image data and learns a unified prompt representation from rich domain type in the domain bank, it can be adapted to more unseen domains in richer open-world scenarios.

\section{Experiments}
Our code is implemented in PyTorch (and will be open source).
The datasets and experiment settings used are described below.

\textbf{Datasets.}
We experiment on 5 benchmark datasets including PACS \cite{PACS} (4 domains, 9,991
samples, 7 classes), VLCS \cite{VLCS} (4 domains, 10,729 samples, 5 classes), OfficeHome \cite{OfficeHome} (4 domains, 15,588 samples, 65 classes), TerraIncognita \cite{Terra} (4 domains, 24,778 samples, 10 classes), and DomainNet \cite{DomainNet} (6 domains, 586,575 samples, 345 classes).
These datasets are all representative and widely used in DG tasks.
In a DG task, experiments are usually performed using the leave-one-out strategy.
For a dataset, one of the domains is selected as the target domain at a time, and the other domains are used as the source domains.
In the SFDG task proposed in this paper, one domain is also selected as the target domain each time, but the data from other domains will not be used as the source domain.

\textbf{Details.}
We adopt the pretraining weights of CLIP and apply ViT-B/16 \cite{ViT} as the backbone for the experiments without special instructions.
More experiments using other backbones can be found in Appendix.2.
The CAE is an autoencoder consisting of linear layers and activation functions, and the number of neurons in the hidden layer is (512, 256, 512).
The optimizer is AdamW with a learning rate of 0.04.
The max epochs is 1000, since the training data is scarce and CAE needs to be finely tuned to achieve a better reconstruction.

\subsection{Domain Generalization}
We compare our approach with some strong DG baselines including SOTA.
The results are shown in Table~\ref{tab:dg}.
As discussed in Section~\ref{sec:related}, some of the compared methods incorporate elaborate learning algorithms
and some works incorporate ensemble learning. It is noteworthy that all these methods require more or less training on the source domain.
Table~\ref{tab:dg} shows that our method achieves over or near SOTA performance on most datasets.
We also notice that our method does not perform well on TerraIncognita.
We speculate that the possible reasons are (1) different domains are simply taken from different cameras, so the domain gaps of TerraIncognita are much smaller compared to other datasets such as PACS; (2) some categories in TerraIncognita are rare (e.g. bobcat) and have small differences between classes (e.g., bobcat and cat), which makes source-free prediction extremely difficult.

\begin{table*}
	\caption{Accuracy (\%) on PACS, VLCS, OfficeHome, TerraIncognita, and DomainNet.
	The last column is the average results of the former five columns. The best results are in \textbf{bold faces} and the second best results are \underline{underlined}. MP and CAE represent the ``mean pooling'' and the ``cosine autoencoder'' strategy in Fig.~\ref{fig:duprg}, respectively.}
	\label{tab:dg}
	\resizebox{\linewidth}{!}{
		\begin{tabular}{l|c|c|c|c|c|c}			
			\toprule
			Method &  PACS &   VLCS &  OfficeHome &  TerraInc. &  DomainNet & Avg. \\
			\midrule
			MMD \cite{Li_2018_CVPR}   & $84.7\pm0.5$ & $77.5\pm0.9$ & $66.3\pm0.1$ & $42.2\pm1.6$ & $ 23.4\pm9.5$ & 58.8\\
			Mixstyle \cite{zhou2021domain}   & $85.2\pm0.3$ & $77.9\pm0.5$ & $60.4\pm0.3$ & $44.0\pm0.7$ & $ 34.0\pm0.1$ & 60.3\\
            GroupDRO \cite{Sagawa2020Distributionally}   & $84.4\pm0.8$ & $76.7\pm0.6$ & $66.0\pm0.7$ & $43.2\pm1.1 $ & $ 33.3\pm0.2$ & 60.7 \\
            IRM \cite{arjovsky2019invariant}   & $83.5\pm0.8$ & $78.5\pm0.5$ & $64.3\pm2.2$ & $47.6\pm0.8 $ & $ 33.9\pm2.8$ & 61.6 \\
            Fish \cite{shi2022gradient}   & $85.5\pm0.3$ & $77.8\pm0.3$ & $68.6\pm0.4$ & $45.1\pm1.3 $ & $ 42.7\pm0.2$ & 63.9 \\
	        ERM \cite{Ishaan2022in} & $84.2\pm0.1$ & $77.3\pm0.1$ & $67.6\pm0.2$ & $47.8\pm0.6$ & $ 44.0\pm0.1$ & 64.2\\
	        SagNet \cite{Nam_2021_CVPR}   & $86.3\pm0.2$ & $77.8\pm0.5$ & $68.1\pm0.1$ & $48.6\pm0.1$ & $ 40.3\pm0.1$ & 64.2\\
	        SelfReg \cite{Kim_2021_ICCV}  & $85.6\pm0.4$ & $77.8\pm0.9$ & $67.9\pm0.7$ & $47.0\pm0.3$ & $ 42.8\pm0.0$ & 64.2\\
	        CORAL \cite{sun2016deep}   & $86.2\pm0.3$ & $78.8\pm0.6$ & $68.7\pm0.3$ & $47.6\pm1.0 $ & $ 41.5\pm0.1$ & 64.5\\
	        mDSDI \cite{NEURIPS2021_b0f2ad44}   & $85.4\pm0.4$ & $79.0\pm0.0$ & $70.5\pm0.4$ & $50.4\pm1.1$ & $ 44.3\pm0.2$ & 65.9\\
	        MIRO \cite{cha2022domain}   & $85.4\pm0.4$ & $79.0\pm0.0$ & $70.5\pm0.4$ & $50.4\pm1.1$ & $ 44.3\pm0.2$ & 65.9\\
	        SWAD \cite{swad} & $88.1\pm0.1$ & $79.1\pm0.1$ & $70.6\pm0.2$ & $50.0\pm0.3$ & $46.5\pm0.1$ & $66.9$ \\
	        DPL \cite{zhang2021amortized} & $\textbf{97.3}\pm0.2$ & $\textbf{84.3}\pm0.4$ & $\textbf{84.2}\pm0.2$ & $\underline{52.6}\pm0.6$ & $-$ & $-$\\
	        GVRT \cite{min2022grounding} & $85.1\pm0.3$ & $79.0\pm0.2$ & $70.1\pm0.1$ & $48.0\pm0.2$ & $44.1\pm0.1$ & $65.2$ \\
	        SEDGE \cite{li2022domain} & $96.1\pm0.0$ & $82.2\pm0.0$ & $80.7\pm0.2$ & $\textbf{56.8}\pm0.3$ & $54.7\pm0.1$ & $\textbf{74.1}$ \\
	        \midrule
            Ours-MP (source free)  & $96.5\pm0.0$ & $83.1\pm0.0$ & $82.5\pm0.0$ & $38.3\pm0.0$ & $\underline{59.2\pm0.0}$ & $71.9$\\
	        Ours-CAE (source free) & $\underline{97.1}\pm0.2$ & $\underline{83.9}\pm0.5$ & $\underline{83.6}\pm0.3$ & $42.0\pm1.3$ & $\textbf{59.6}\pm0.3$ & $\underline{73.2}$ \\
	
			\bottomrule
	\end{tabular}
	}
\end{table*}
\subsection{Ablations}

\begin{figure*}[t]
  \centering
  	\subfigure[PACS]{
		\centering
		\includegraphics[width=0.9\columnwidth]{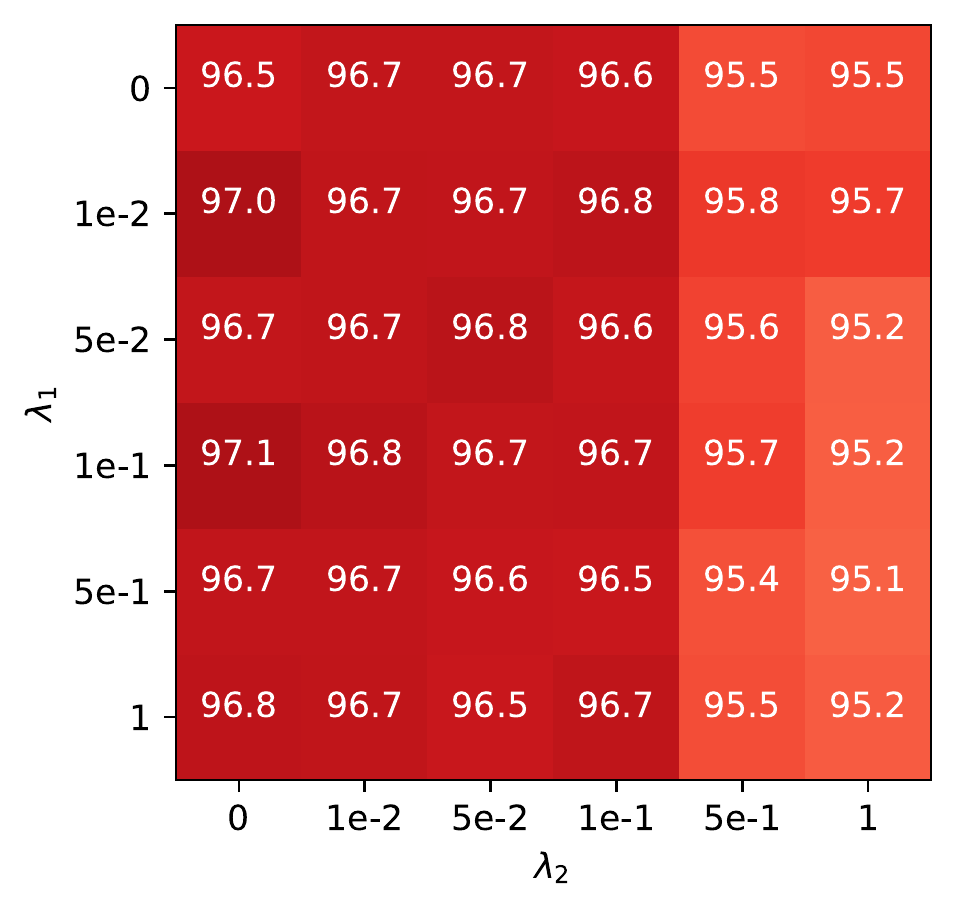}
		\label{fig:mnistgen}
	}
	\subfigure[VLCS]{
		\centering
		\includegraphics[width=0.9\columnwidth]{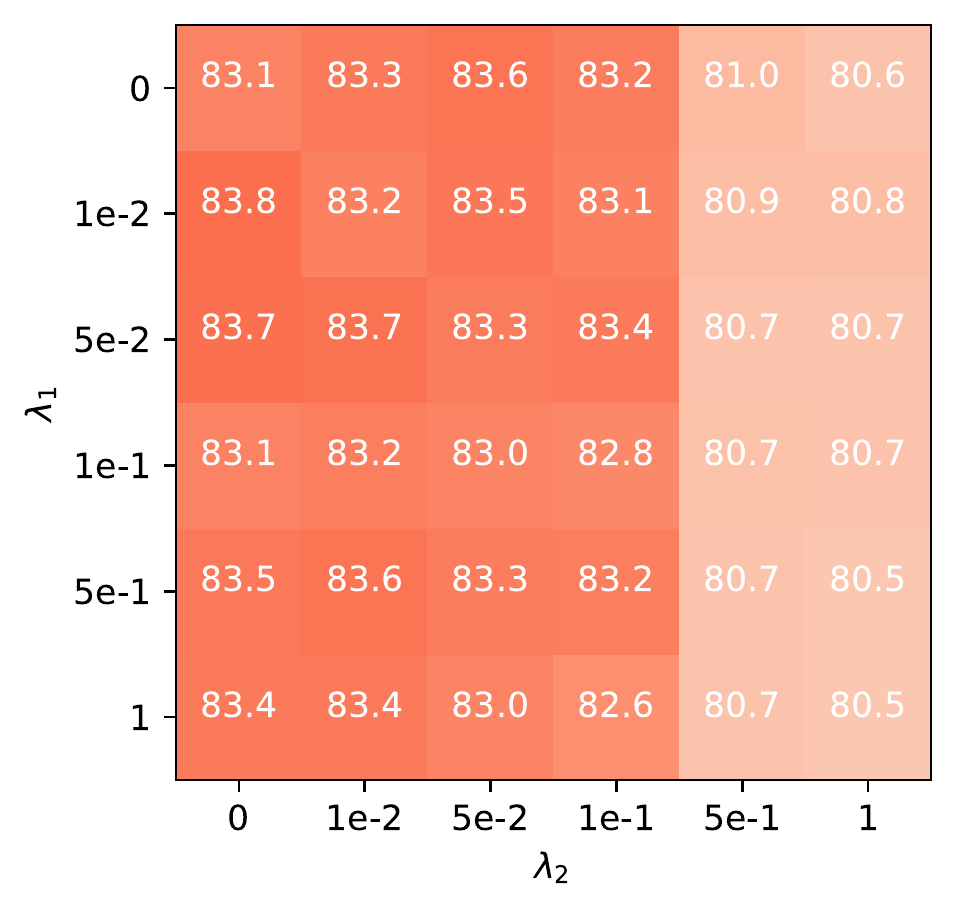}
		\label{fig:realgen}
	}
  \caption{Average accuracy on PACS and VLCS datasets with different combinations of weighting coefficients $[\lambda_1, \lambda_2]$.
  }
  \label{fig:lambda}
\end{figure*}

\textbf{Hyper-parameters.}
Our method introduces two additional hyperparameters (i.e.,the loss weight coefficients $\lambda_1$ and $\lambda_2$ in Eq~\ref{6}).
We tested different combinations of $\lambda_1$ and $\lambda_2$ on PACS and VLCS datasets, which encompassed the ablation experiment (when $\lambda_1=\lambda_2=0$).
The results are shown in Fig.~\ref{fig:lambda}.
We only show the experimental results on the VLCS dataset in order to provide a clearer picture of the effect of hyperparameter variation on the results.
It can be seen that when both loss functions have a performance improvement. In addition, most of the hyperparameter combinations can achieve good performance, and the choice of weights does not require particularly tedious adjustment. Specially, $\lambda_1=\lambda_2=0$ refers to the mean pooling strategy as shown in Fig.~\ref{fig:duprg}

\textbf{Loss for reconstruction.} Inspired by the inference process of CLIP, we choose the cosine similarity as the reconstruction loss $\mathscr{L}_{rec}$ to optimize the autoencoder. The effects of using traditional L2 loss is also given in Table~\ref{tab:loss} for a comparison.

\begin{table*}
	\caption{Accuracy (\%) on SFDG tasks corresponding to different loss functions. ``SP'' represents using standard prompt (``a photo of a \textbf{class}'') for inference. The best results are in \textbf{bold faces}.
	}
	\label{tab:loss}
	\resizebox{\linewidth}{!}{
		\begin{tabular}{l|cccc|c|cccc|c}			
			\toprule
			\multirow{2}{*}{} & \multicolumn{5}{c|}{PACS} & \multicolumn{5}{c}{VLCS}\\
			\cline{2-11}
			 &  Photo &   Art &  Cartoon &  Sketch & Avg. & VOC2007 & LabelMe & Caltech101 & SUN09 & Avg.\\
			\midrule
			SP  & $99.8$ & $97.2$ & $99.1$ & $88.1$ & $96.1$ & $86.0$ & $70.2$ & $99.9$ & $73.6$ & $82.4$ \\
			\midrule
			L2 loss  & $83.0$ & $97.3$ & $67.9$ & $69.1$ & $79.3$ & $72.8$ & $\textbf{72.8}$ & $69.3$ & $69.1$ & $71.0$ \\
			Cosine loss  & $\textbf{99.9}$ & $\textbf{97.8}$ & $\textbf{99.2}$ & $\textbf{91.4}$ & $\textbf{97.1}$ & $\textbf{87.0}$ & $72.5$ & $\textbf{99.9}$ & $\textbf{76.1}$ & $\textbf{83.9}$ \\
	
			\bottomrule
	\end{tabular}}
\end{table*}

\begin{table}[t]
	\caption{Accuracy (\%) on SFDG tasks corresponding to different domain banks.
	An ``empty'' bank means using standard prompt only;
	``task'' means to form the bank with domain type of the corresponding dataset;
	``combined'' represents a summary of the domains of several datasets; and
	``expanded'' adds some additional propmts based on ``combined''.
	The domains of VLCS and TerraIncognita do not correspond to a specific style, so no specific prompt is provided. The best results are in \textbf{bold faces} and the second best results are \underline{underlined}.
	}
	\label{tab:bank}
	\resizebox{\linewidth}{!}{
		\begin{tabular}{l|c|c|c|c|c|c}			
			\toprule
			Bank &  PACS &   VLCS &  OfficeHome &  TerraInc. &  DomainNet & Avg. \\
			\midrule
			Empty  & $96.1$ & $82.4$ & $81.6$ & $36.9$ & $ 56.7$ & $70.7$\\
			Task  & $96.7$ & $82.4$ & $\textbf{83.8}$ & $36.9$ & $57.2$ & $71.4$ \\
			Combined  & $\textbf{97.1}$ & $\textbf{83.9}$ & $\underline{83.6}$ & $\textbf{42.0}$ & $\textbf{59.6}$ & $\textbf{73.2}$ \\
			Expanded & $\underline{96.8}$ & $\underline{82.9}$ & $83.3$ & $\underline{40.8}$ & $ \underline{59.1}$ & $\underline{72.6}$\\
	
			\bottomrule
	\end{tabular}
	}
\end{table}
\textbf{Domain bank.}
The rich and diverse prompt can be considered as the raw material for constructing domain-unified prompt representations.
Compared to collecting domain-specific image datasets, generating prompts in textual form can be very easy.
We integrated the domains of all the used datasets into a domain bank as the defalut setting for all the experiments in this paper, which corresponds to ``combined'' in Table~\ref{tab:bank}. In fact, the domain bank can be easily expanded or reduced.
Table~\ref{tab:bank} shows the experimental results corresponding to different domain banks.
On one hand, when the bank shrinks to the limit (empty set), the prompt set degenerates to the standard prompt.
On the other hand, an overly redundant domain bank can increase the amount of computation while not necessarily improving performance, but may have a better performance for the open-world scenarios.


\begin{figure*}[t]
  \centering
  	\subfigure[PACS]{
		\centering
		\includegraphics[width=0.65\columnwidth]{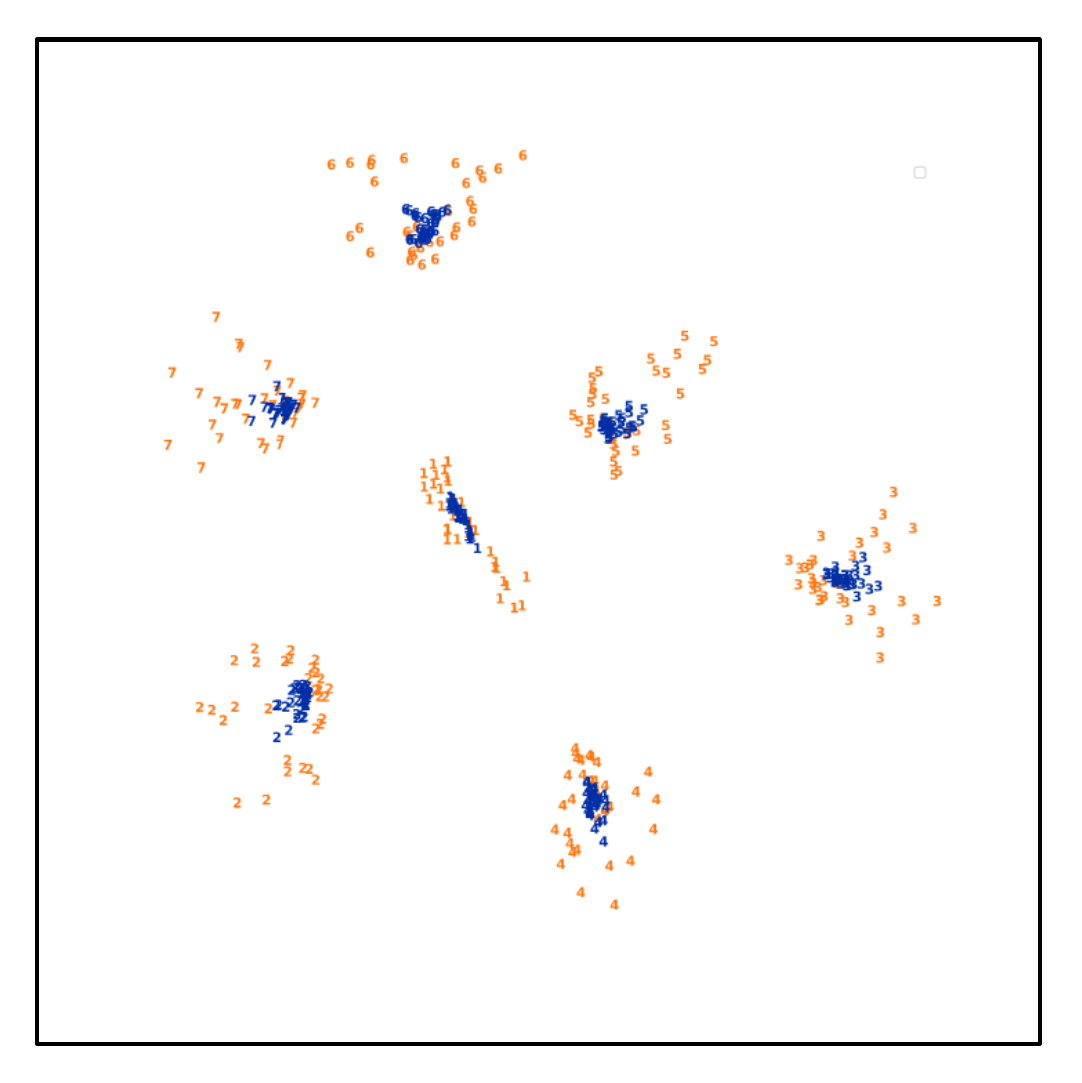}
		\label{fig:mnistgen}
	}
	\subfigure[OfficeHome]{
		\centering
		\includegraphics[width=1.14\columnwidth]{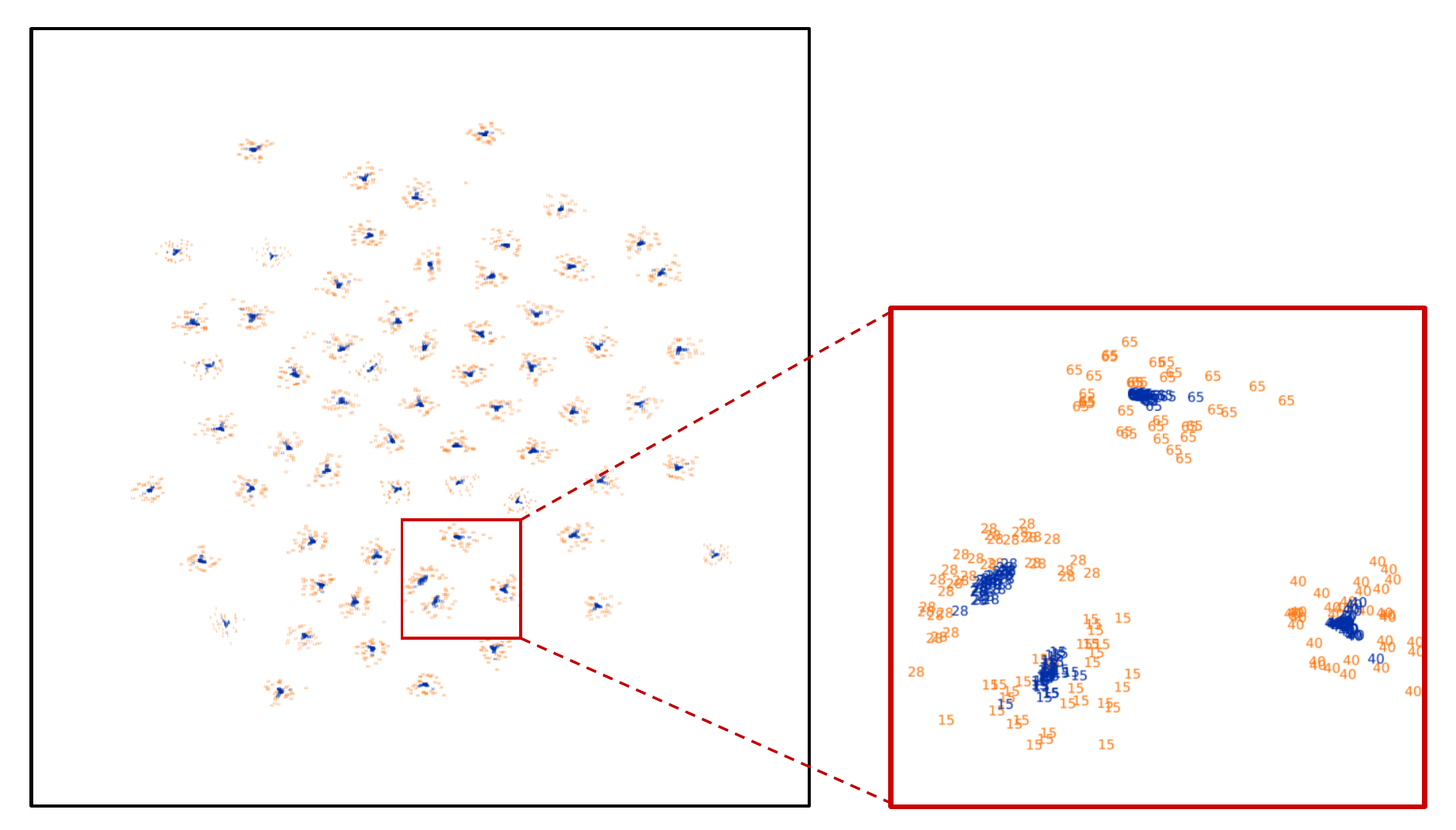}
		\label{fig:realgen}
	}
  \caption{Visualization of the input and output prompt feature distribution of CAE by t-SNE. The numbers represent the categories. The orange and blue points refer to the input and output prompt representations of CAE, respectively. The points with the same color and number stand for the prompt representations generated by various domain types from the domain bank.}
  \label{fig:tsne}
\end{figure*}

\begin{table}[t]
	\caption{Accuracy (\%) on domains in open-world scenarios. ``X'', ``G'', ``P'', ``A'', ``C'', and ``S'' represent the pixelate and geometric, photo, art, cartoon, and sketch style, respectively. The best results are in \textbf{bold faces}.}
	\label{tab:openset}
	\resizebox{\linewidth}{!}{
		\begin{tabular}{l|cc|cc|cc|cc}			
			\toprule
			\multicolumn{9}{l}{Multi-Source Methods} \\
			\midrule
			 & \multicolumn{2}{c|}{P A C}& \multicolumn{2}{c|}{P A S}& \multicolumn{2}{c|}{P C S}& \multicolumn{2}{c}{A C S} \\
			 \midrule
			 &  X &   G &X &   G &X &   G &X &   G  \\
			\midrule
			ERM \cite{Ishaan2022in} & 51.5 & 61.5 & 48.2 & 65.1 & 57.9 & 58.9 & 53.8 & 62.1 \\
			GroupDRO \cite{Sagawa2020Distributionally} & 52.2 & 61.3 & 51.6 & 62.9 & 60.4 & 59.7 & 55.3 & 64.1 \\
			CORAL \cite{sun2016deep} & 46.1 & 57.1 & 45.0 & 62.2 & 51.1 & 57.2 & 49.2 & 63.5\\
			DPL \cite{zhang2021amortized} & 94.9 & 90.8 & 91.9 & 91.3 & 96.0 & 93.8 & 93.7 & 92.9 \\
			\midrule
			\multicolumn{9}{l}{Source-Free Methods} \\
			\midrule
			 & \multicolumn{4}{c|}{X} & \multicolumn{4}{c}{G} \\
			 \midrule
			Ours (CAE) & \multicolumn{4}{c|}{\textbf{97.1}} & \multicolumn{4}{c}{\textbf{95.1}} \\
	
			\bottomrule
	\end{tabular}
	}
\end{table}

\subsection{Visualizations}
To further demonstrate the effect of the proposed CAE on generating domain-unified prompt representations, we visualize the input and output feature distributions of CAE by t-SNE \cite{van2008visualizing} on PACS and OfficeHome datasets. As shown in Fig.~\ref{fig:tsne}, the input prompt representations of different domains (orange) are relatively scattered, while the output prompt representations (blue) are closely distributed.
This means that out CAE filter out the domain-related part while the class-related part is preserved.

Since both strategies calculate the domain-unified prompt representations by utilizing the mean pooling operation in Fig.~\ref{fig:duprg}, the center of the prompt representations serves as the domain-unified prompt representations for each class. As shown in the zoomed-in part of Fig.~\ref{fig:tsne}(b), the center of the output features (blue) has a clear shift compared to that of the input features (orange), which is brought by is the filtering of domain-related parts.
Therefore, our CAE can be used to generate better domain-unified prompt representations.

\subsection{Open-world Domain Generalization}\label{Open-world}
To demonstrate the advantages of our approach on the domain generalization task in open-world scenarios, we collected data of two domains (i.e., pixelate style and geometric style) that hardly appear in the current DG datasets from internet as a supplement to the PACS dataset (share the same categories with the PACS dataset). More details about the datasets can be found in Appendix.1. As shown in Table~\ref{tab:openset}, our method that need no source domain data for training significantly outperforms recent DG methods by a large margin.
\section{Conclusion}
In this work, we develop a more challenging version of the domain generalization problem that requires good performance in the target domain without using the source domain data.
This paper also proposes a feasible solution to the source-free domain generalization problem.
By transforming the need for image feature consistency into constructing domain-unified prompt representations through a vision-language model, we bypass the need for source domain data. Experimental results demonstrate that our method exhibits excellent performance on most of the tested datasets without using source domain data, achieving a competitive performance compared to other approaches trained with source domain data. More importantly, we substantially outperformed current SOTA DG methods on open-world domains without training with multi-source domain data.


%





\ifCLASSOPTIONcaptionsoff
  \newpage
\fi



%

\bibliographystyle{IEEEtran}
\bibliography{bare_jrnl_compsoc}
\newpage
\appendix

\section{Appendix}

\subsection{Openset Datasets}\label{app:dataset}
\begin{figure*}[t]
  \centering
  \includegraphics[width=\linewidth]{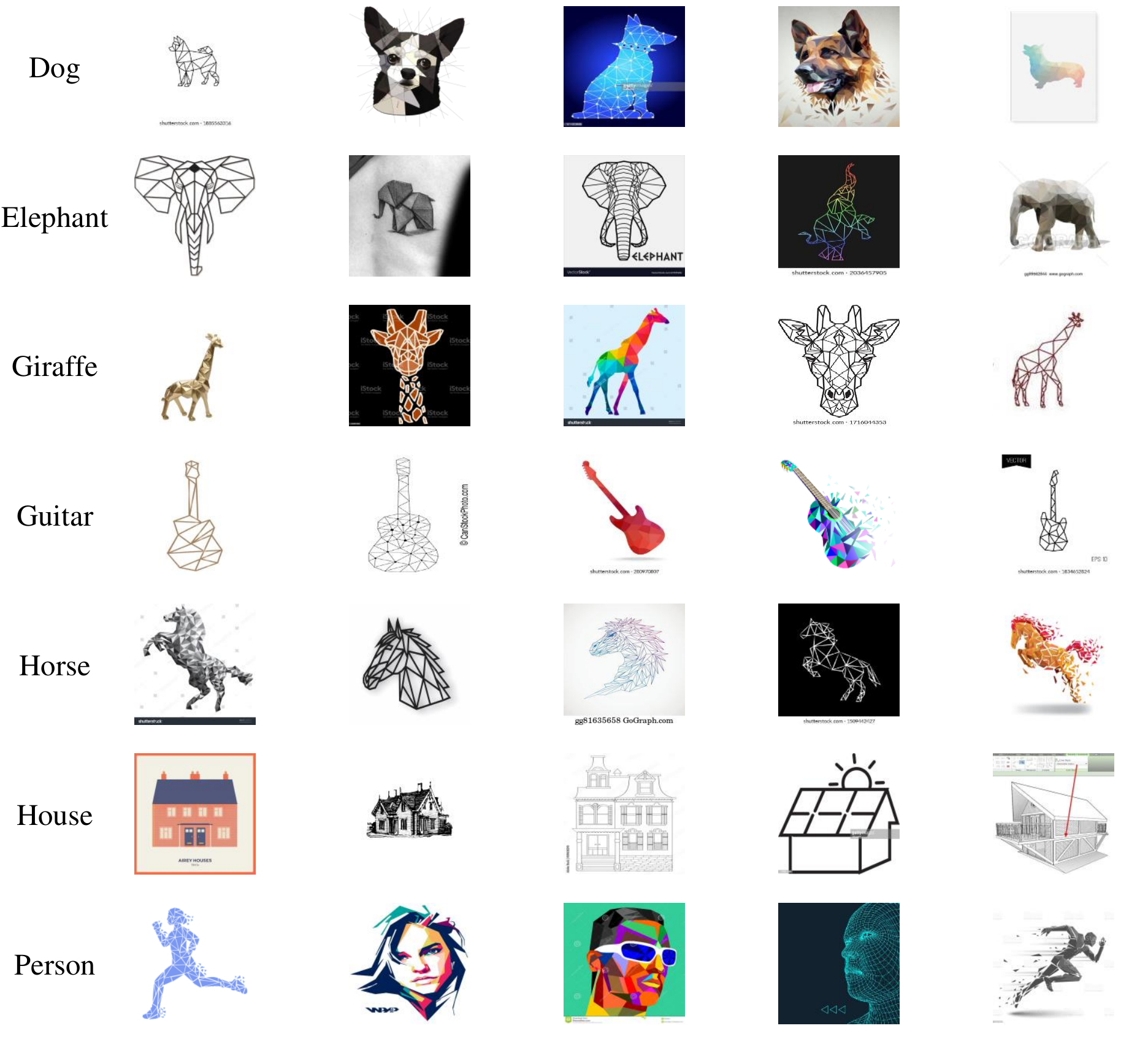}
  \caption{Geometric style images in our dataset}
  \label{fig:geometric}
\end{figure*}

\begin{figure*}[t]
  \centering
  \includegraphics[width=\linewidth]{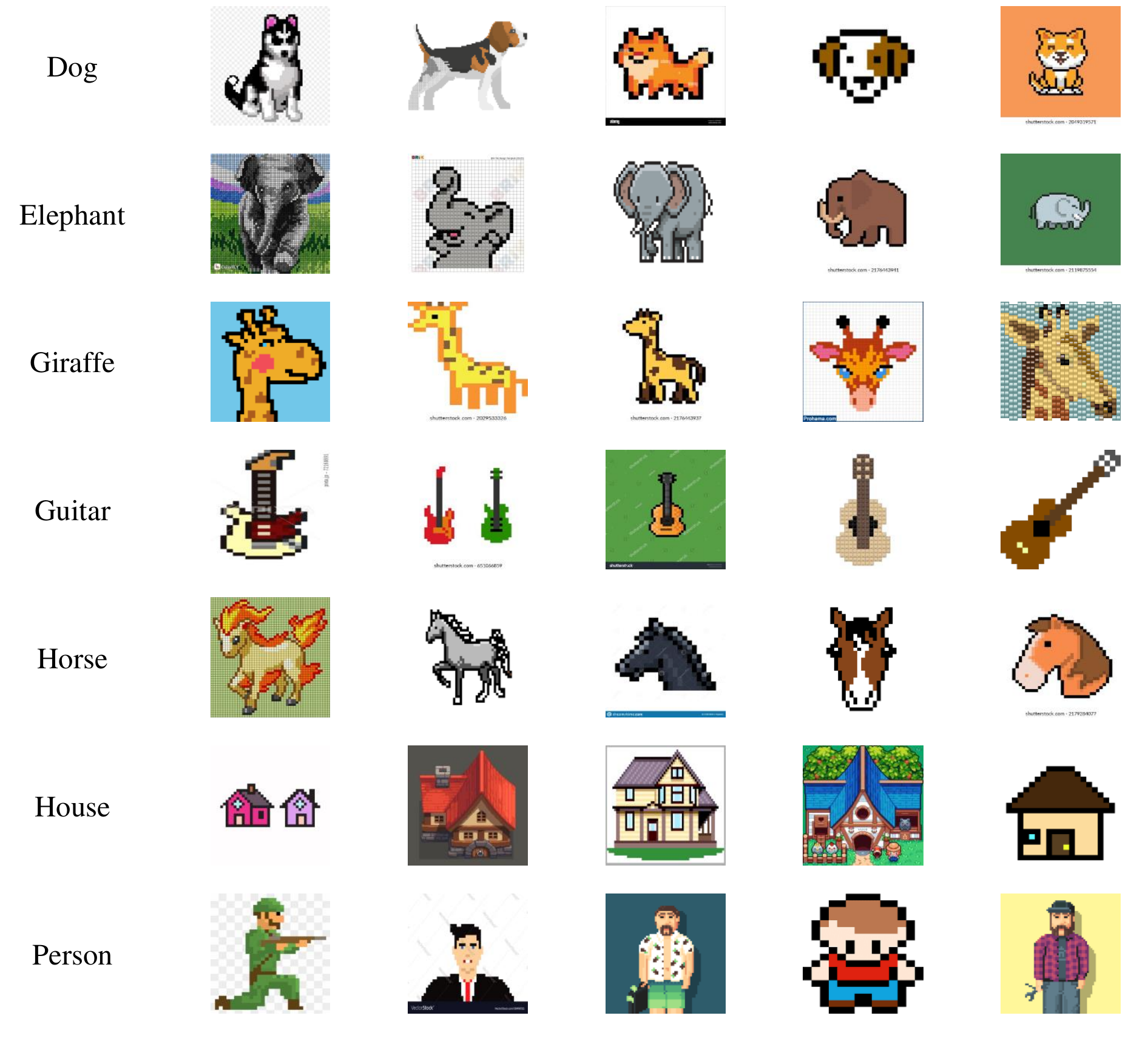}
  \caption{Pixelate style images in our dataset}
  \label{fig:pixel}
\end{figure*}

As declared in Section~\ref{Open-world}, we collect a supplementary for PACS datasets to evaluate the effectiveness of our method on the domain generalization task in open-world scenarios. Specifically, we collect 1365 pixelate style images and 1202 geometric style images from the internet. As shown in Fig.\ref{fig:geometric} and Fig.\ref{fig:pixel}, some samples in the datasets are given. The datasets will be released together with our source code.

\subsection{Backbones}\label{app:backbone}

In the proposed framework, the only part that needs to be trained is the proposed CAE (domain-unified prompt representation generator), while the parameters of the text and image encoders are fixed.
These encoders are replaced to verify the generalizability of the proposed approach for different model structures.
We implemented ResNet \cite{ResNet} and ViT \cite{ViT} and generate prompt representations with different strategies,
\textbf{``SP''}: using standard prompt ("a photo of a \textbf{\{class\}}") to generate the only representation for each category;
\textbf{``MP''}: the prompts generated by different domains are aggregated by the mean pooling operation;
\textbf{``CAE''}: using CAE to generate the unified representations.
The results are shown in Table~\ref{tab:backbone}.
Regardless of the model structure, using CAE to generate unified-representations is always the best for both ViT-based backbones and is comparable with the mean pooling strategy for ResNet50, which demonstrate the effectiveness of the proposed CAE.


\begin{table*}[t]
	\caption{Accuracy (\%) on SFDG tasks corresponding to different encoder backbones. Bold denotes the best.
    }
	\label{tab:backbone}
	\resizebox{\linewidth}{!}{
		\begin{tabular}{l|c|c|c|c|c|c}			
			\toprule
			Backbone &  PACS &   VLCS &  OfficeHome &  TerraInc. &  DomainNet & Avg. \\
			\midrule
			ResNet50 (SP)  & $91.1$ & $80.8$ & $71.3$ & $17.8$ & $ 45.6$ & $61.3$\\
			ResNet50 (MP)  & $93.0$ & $\textbf{82.5}$ & $\textbf{75.3}$ & $24.6$ & $ 48.1$ & $\textbf{64.7}$\\
			ResNet50 (CAE)  & $\textbf{93.1}$ & $81.2$ & $73.0$ & $\textbf{25.2}$ & $ \textbf{48.4}$ & $64.2$\\
			\midrule
			ViT-B/16 (SP)  & $96.1$ & $82.4$ & $81.6$ & $36.9$ & $ 56.7$ & $70.7$\\
			ViT-B/16 (MP)  & $96.5$ & $83.1$ & $82.5$ & $38.3$ & $59.2$ & $71.9$\\
			ViT-B/16 (CAE)  & $\textbf{97.1}$ & $\textbf{83.9}$ & $\textbf{83.6}$ & $\textbf{42.0}$ & $\textbf{59.6}$ & $\textbf{73.2}$ \\
			\midrule
			ViT-B/32 (SP) & $94.7$& $81.3$ & $78.8$ & $23.2$ & $53.2$ & $66.2$\\
			ViT-B/32 (MP) & $94.7$ & $81.1$ & $79.6$ & $26.4$ & $55.0$ & $67.4$\\
			ViT-B/32 (CAE) & $\textbf{95.1}$ & $\textbf{81.9}$ & $\textbf{80.3}$ & $\textbf{39.9}$ & $\textbf{55.7}$ & $\textbf{70.6}$\\
	
			\bottomrule
	\end{tabular}
	}
\end{table*}

\end{document}